# Graph-Based Deep Learning on Stereo EEG for Predicting Seizure Freedom in Epilepsy Patients

Artur Agaronyan[1], Syeda Abeera Amir[1], Nunthasiri Wittayanakorn[2], John Schreiber[3], Marius G. Linguraru[4], William Gaillard[3], Chima Oluigbo[2], Syed Muhammad Anwar[4]

*Abstract*— Predicting seizure freedom is essential for tailoring epilepsy treatment. But accurate prediction remains challenging with traditional methods, especially with diverse patient populations. This study developed a deep learning-based graph neural network (GNN) model to predict seizure freedom from stereo electroencephalography (sEEG) data in patients with refractory epilepsy. We utilized high-quality sEEG data from 15 pediatric patients to train a deep learning model that can accurately predict seizure freedom outcomes and advance understanding of brain connectivity at the seizure onset zone. Our model integrates local and global connectivity using graph convolutions with multi-scale attention mechanisms to capture connections between difficult-to-study regions such as the thalamus and motor regions. The model achieved an accuracy of 92.4% in binary class analysis, 86.6% in patient-wise analysis, and 81.4% in multi-class analysis. Node and edge-level feature analysis highlighted the anterior cingulate and frontal pole regions as key contributors to seizure freedom outcomes. The nodes identified by our model were also more likely to coincide with seizure onset zones. Our findings underscore the potential of new connectivity-based deep learning models such as GNNs for enhancing the prediction of seizure freedom, predicting seizure onset zones, connectivity analysis of the brain during seizure, as well as informing AI-assisted personalized epilepsy treatment planning.

## I. INTRODUCTION

Epilepsy affects approximately 50 million people worldwide, with nearly one-third experiencing drug-resistant epilepsy (DRE), where seizures persist despite multiple anti-seizure medications [1]. For these patients, surgical intervention is often necessary to achieve full or partial seizure freedom. Surgical resections can be extensive however, based on the network of brain regions propagating the seizure. There is strong evidence that surgery can be highly effective in achieving seizure freedom and improving quality of life in patients with drug-resistant-epilepsy [2], [3]. Furthermore, the World Health Organization [1] estimates that up to 70% of individuals with epilepsy could live seizure-free if properly diagnosed and treated, including patients with DRE.

The success of epilepsy surgery heavily depends on accurate identification of the epileptogenic network and prediction of post-surgical outcomes [4]. This critical need for accurate seizure freedom prediction has driven research into developing more sophisticated analytical approaches to solve this issue [4], [5]. Stereo electroencephalography (sEEG) and subdural electrode (SDE) implantation are among the most widely used invasive monitoring methods for identifying the epileptogenic network, with significant evidence supporting better seizure freedom outcomes when utilizing sEEG [2], [6]. sEEG offers deep brain recording capabilities with high temporal resolution and is often used for the identification of discrete cortical seizure onset zones (SOZs) in surgical planning. Among the subcortical regions involved in propagation of seizures, thalamic nuclei have repeatedly shown interconnectedness with ictal brain regions. In a study investigating energy distribution between temporal cortices across seizure stages, the average thalamic power was found to be significantly higher at seizure onset compared to baseline power [2].

The increasing inclusion of thalamic recordings in sEEG implementations has opened new avenues for understanding thalamocortical networks, which are fundamental to both developing brain function and pathological states [7]. Despite this wealth of data, traditional analysis methods often struggle to capture the complex, interconnected nature of epileptic networks, particularly the subtle patterns that may predict treatment outcomes. Graph Neural Networks (GNNs) present a promising approach for analyzing such complex neural data, as they can model the brain's networked structure and capture both local and global connectivity patterns [8], [9]. In graph-based applications brain regions are represented as nodes and the strength of their connection as edges, hence enabling the representation of connectivity patterns. Therefore, unlike traditional machine learning approaches, GNNs can explicitly incorporate spatial relationships and non-linear interactions between brain regions, making them particularly well-suited for analyzing thalamocortical connectivity patterns in epilepsy.

In this paper, we present a GNN-based classifier model for predicting seizure freedom outcomes using sEEG data. Our method combines spatial mapping from GNNs with data offering high temporal precision to capture the complex interactions between various brain regions, especially the critical connections between the thalamus and cortical structures. We analyzed classification of post-resection seizure freedom per patient as well as per individual seizure. We also applied our model to study connectivity during seizures and identify the most important regions for classification. This approach not only improves prediction accuracy but also provides valuable insights into poorly understood seizure dynamics, such as the relationship between seizure onset zones and seizure networks.

[1]Artur Agaronyan and Syeda Abeera Amir are with Sheikh Zayed Institute for Pediatric Surgical Innovation, Children's National Hospital, Washington, DC

[2]Nunthasiri Wittayanakorn and Chima Oluigbo are with the Department of Neurosurgery, Children's National Hospital, Washington DC

[3]John Schreiber and William Gaillard are with Pediatric Neurology, Children's National Hospital, Washington DC

[4]Syed Muhammad Anwar and Marius G. Linguraru are with Children's National Hospital, Washington DC and School of Medicine and Health Sciences, George Washington University, Washington DC (sanwar@childrensnational.org)

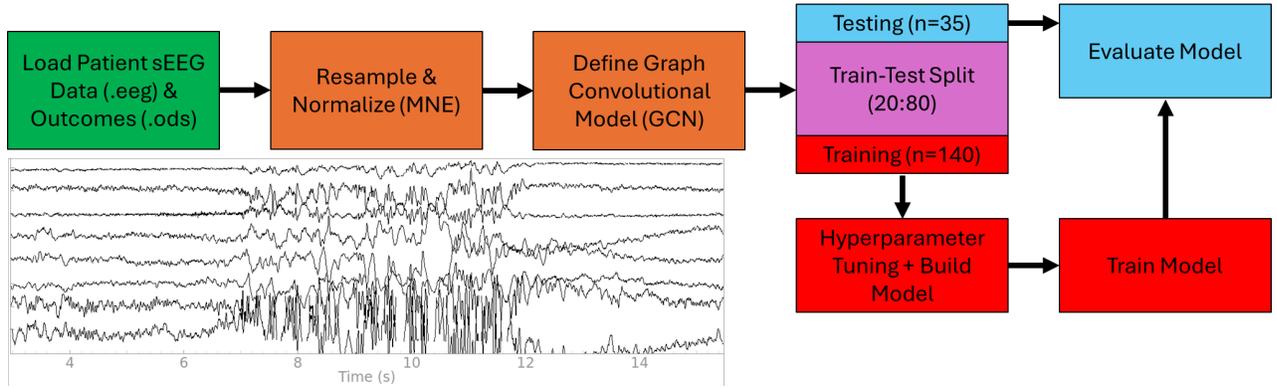

**Figure 1:** *Overview of the proposed pipeline to predict seizure freedom using sEEG data.*

## II. PROPOSED METHODOLOGY

### A. Patient Selection and Implantation

Fifteen pediatric patients with pharmaco-resistant epilepsy underwent stereo electroencephalography (sEEG) with integrated thalamic subdivision interrogation at Children's National Hospital in Washington, DC. This study received approval from the Children's National Hospital Institutional Review Board. The sEEG electrodes targeting thalamic nuclei were selected based on the suspected epileptogenic region's location and the known anatomical connectivity of thalamic nuclei. Postoperative CT scans were used to segment and reconstruct each depth electrode targeting a thalamic nucleus, utilizing ROSA ONE Brain and Surgical Theater for accuracy, particularly in identifying the number of contacts successfully placed within each nucleus. Any electrode contacts located outside the planned thalamic subdivision were excluded from the analysis. Each patient was later rated by the clinical team with a four-class Engel score [10] for seizure freedom outcome. Table 1 shows the distribution of patients.

**Table 1.** Distribution of seizure classes by patient.

| Class (Engel) | Patients | Seizures | | Channels |
|---|---|---|---|---|
| I | 1 | 10 | 10 | 95 |
| II | 2 | 28 | 14 | 95 |
| | | | 4 | 80 |
| III | 8 | 100 | 15 | 84 |
| | | | 24 | 95 |
| | | | 10 | 100 |
| | | | 4 | 95 |
| | | | 7 | 69 |
| | | | 8 | 83 |
| | | | 23 | 78 |
| | | | 9 | 87 |
| IV | 4 | 47 | 3 | 58 |
| | | | 8 | 127 |
| | | | 22 | 127 |
| | | | 14 | 79 |
| Total | 15 | 175 | | 1352 |

### B. Model Development and Testing

Engel scores were first converted into classes. For the binary model: scores I and II were treated as a positive outcome, and scores III and IV were treated as a negative outcome. For the multi-class model: Engel score I was class I, score II was class II, and scores III and IV were treated as class III. The preprocessing of raw EEG data involved several steps utilizing MNE-Python [11]. EEG data was resampled to a fixed sampling rate of 128 Hz. Each EEG sample was normalized by subtracting the mean and dividing by the standard deviation of the signal. Subsequently, the samples were either padded or truncated to a fixed length of 5000 time points to ensure uniformity in input dimensions.

Figure 1 shows the overall flow of the pipeline. Following preprocessing, all EEG data samples had consistent shapes. This processed EEG data was used to create graph data structures. Each channel was treated as a node, and edges were established between all pairs of nodes based on correlation between channels. The data was split into training and test sets with stratified sampling to ensure both classes were adequately represented in each set.

The GNN architecture consisted of two graph convolutional layers, followed by a global mean pooling layer to aggregate node features. Rectified linear unit (ReLU) activation functions were applied after each convolutional layer. The model's output was a two-dimensional or three-dimensional vector, corresponding to the binary and multi-class approaches. We used the cross-entropy loss function and the Adam optimizer for training the model. Hyperparameter tuning was performed using Optuna [12]. The study was run for 100 epochs. Training time on a single Nvidia RTX A5000 GPU was about one hour for all 175 samples.

Feature analysis of the model was performed, and the top 10 most important channels were identified. The model was then re-run for the 3-class purpose with only these channels. The model's performance was evaluated by accuracy, precision, recall, and the F1 score. The data was then analyzed for correlation between channels by Pearson Correlation given as

$$p = \frac{Cov(X,Y)}{\sigma_X \sigma_Y} \quad , (1)$$

where X and Y are the channel pairs analyzed for correlation. The correlations were visualized in a graph network to compute thalamic nodes' eigenvector centrality

$$x_v = \frac{1}{\lambda} \sum_{t \in G} a_{v,t} x_t \quad , \quad (2)$$

where $a_{v,t}$ is the adjacency matrix, $x_t$ is the eigen vector of a, $\lambda$ is the largest eigen vector, and G is the set of vector neighbors; and network density as follows

$$\rho = \frac{2E}{n(n-1)} \quad , \quad (3)$$

where $E$ is the number of connections (edges) in the network and $n$ is the total number of nodes.

## III. EXPERIMENTAL RESULTS

Our model achieved an accuracy of 86.6% for patient-wise analysis. Precision was 85.1%, recall was 80.56% and F1 score was 82.3%. For binary classification, accuracy was 91.4%, precision was 91.5%, recall was 91.4%, and F1 score was 91.2%. For multi-class analysis, accuracy was 81.4%, precision was 81.1%, and recall was 81.4%. F1 score was 80.9%. The confusion matrix showed that for class I, the model correctly predicted 13 out of 19 instances, misclassifying 4 as class II and 2 as class III. For class II, the model correctly predicted 37 out of 40 instances, misclassifying 2 as class I and 1 as class III. For class III, the model correctly predicted 7 out of 11 instances, with 1 misclassified as class I and 3 as class II.

Feature analysis revealed that specific brain regions contributed more significantly to the model's conclusions. Figure 2 shows the regions identified as most important, such as the anterior cingulate, thalamus, and anterior frontal pole, for predicting seizure freedom. When the model was run utilizing only the top 10 most important channels, accuracy was 71.2%. Table 2 shows a summary of the metrics generated. Figure 3 shows the connectivity in the thalamocortical network engaged during a seizure with and without seizure freedom post-resection. The network graphs show stronger correlation between cortical and thalamic channels in the patient without seizure freedom, with an average thalamic node connection strength of 5.340, compared to 1.442 of patient with seizure freedom. Overall network A has a denser network ($\rho$ = 0.533) compared to B ($\rho$ = 0.3), with the thalamic node's eigenvector centrality being similar (A = 0.506, B = 0.582) for both.

**Table 2.** Results from our pipeline predicting seizure freedom.

|  | 3-Class | 2-Class | Patient-Wise | 3-Class w/ Top 10 Channels |
|---|---|---|---|---|
| Accuracy | 81.4% | 91.4% | 86.6% | 71.2% |
| Precision | 81.1% | 91.5% | 85.1% | 70.1% |
| Recall | 81.4% | 91.4% | 80.5% | 71.2% |
| F1 | 80.9% | 91.2% | 82.3% | 71.0% |

## IV. DISCUSSION

Our model achieved high accuracy with all three approaches, with 86.6% accuracy in patient-wise analysis, 81.4% accuracy in multi-class analysis, and an impressive 91.4% accuracy in binary classification. The anterior cingulate, thalamus, and frontal pole regions identified in our feature analysis as important matched the SOZs identified by the traditional sEEG analysis. The high ratings for the thalamic (8.65) and insular (8.54) regions in particular highlight the interplay between those regions' contributions to seizure freedom outcome. Connectivity analysis further identified that the patient without seizure freedom had a denser network between the thalamus and SOZ. The patient with seizure freedom, however, had a less strong correlation, suggesting that excessive neural activity between the thalamus and SOZ may contribute to a lack of seizure freedom. This may explain why the model picked those regions as important.

Predicting the outcome pre-surgery can help clinicians decide on the most effective course of treatment to achieve seizure freedom. Our analysis explored the impact of training with only the top 10 most important channels. While the model achieved an accuracy of 81.4% in the full 3-class classification, this dropped to 71.2% when limited to the top 10 channels. Limiting the model to only the most important channels, while potentially simplifying the model and reducing computational time, also sacrifices valuable information present in the full dataset. However, even with only the top 10 channels, the model achieved reasonable performance.

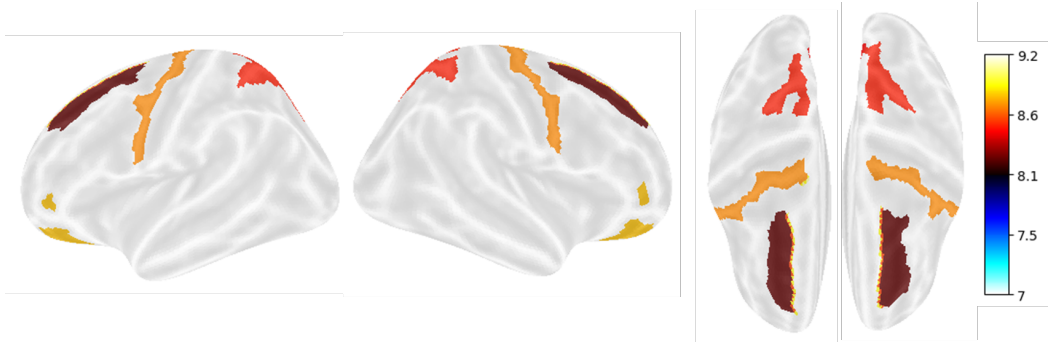

**Figure 2:** Regions for predicting seizure freedom based on feature significance mapped onto the brain: Mid-Frontal-Central (Motor)-Superficial Frontal-Fronto-Polar (9.29), Caudal Lateral-Orbitofrontal region (8.81), Superficial Frontocentral-Anterior-Temporal (8.65), Mid-Frontal-Central (Motor) (8.72), Posterior-Medial-Parietal-Insular (8.54), Superior Central-Midline-Insular (8.50), and Superficial Frontal-Fronto-Polar (8.24) regions.

One notable characteristic of our study is the relatively small sample size with which the model was able to attain high accuracy, with only 15 patients. This may indicate that a connectivity-based deep learning approach such as a GNN may be more useful than traditional machine learning methods for this purpose, particularly for assisting with treatment of specialized and diverse patient populations, where patient populations may be limited.

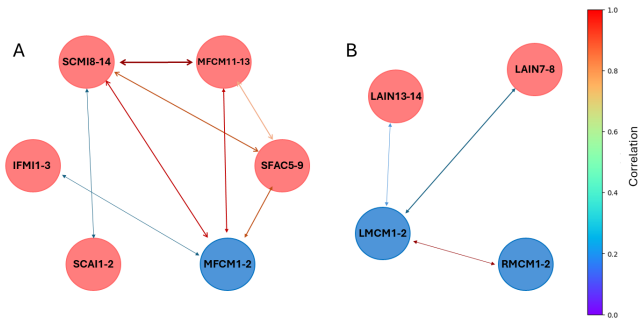

**Figure 3.** Connectivity analysis for two patient groups; **(A)** non-seizure free **(B)** seizure free. Blue nodes represent thalamic channels, and the red nodes represent cortical SOZ channels. SCM: Superior Central-Midline-Insular, MFCM: Mid-Frontal-Central-Motor, IFMI: Inferior Frontal-Anterior-Insular, SCAI: Superior Central-Anterior-Insular, SFAC: Superficial Frontocentral-Anterior-Temporal, LMCM/RMCM: Left/Right Centro-median Thalamic Nucleus, LAIN: Left Anterior Insular.

Utilization of sEEG data in conjunction with advanced deep learning techniques like we present in this study offers a highly promising direction for personalized seizure treatment planning. Enhancing model accuracy through the incorporation of additional modalities such as MRI and PET could also provide a more comprehensive understanding of epilepsy. However, variations in individual patient characteristics, such as the type and location of epilepsy, may not be fully captured in a small sample size such as ours. Future studies that include larger patient populations can refine the model's performance, ensuring its robustness and reliability in different clinical settings. Expanding the sample size would also enable the exploration of more nuanced patterns and relationships within the data, ultimately enhancing the model's predictive capabilities and clinical utility.

## V. CONCLUSION

This study presents a pioneering pipeline for predicting seizure freedom and understanding seizure dynamics in patients with refractory epilepsy using sEEG. Our results demonstrate high accuracy for multi-class analysis, binary-class analysis, as well as patient-wise analysis, with a relatively small amount of training data. Our finding that the channels marked as important by the model correlated with the channels manually identified as SOZs demonstrate the potential of GNNs functioning as part of an AI-assisted pipeline for guided epilepsy treatment, helping inform clinicians about personalized seizure characteristics before any treatment has occurred. The implications of this work extend beyond predicting seizure freedom and SOZs; GNNs such as the one we developed may help guide future research to understand the connectivity of the brain and how the brain reacts to disruptions in its connectivity networks during traumatic neurological events such as seizures, traumatic brain injury, and diffuse axonal injury.